\documentclass[runningheads]{llncs}
\usepackage{lineno,hyperref}
\usepackage[export]{adjustbox}
\usepackage{amssymb}
\usepackage{multicol}
\usepackage{graphicx} 
\usepackage{filecontents}
\usepackage{bbold}
\usepackage{makeidx}  
\usepackage[misc]{ifsym}
\usepackage{amsmath,graphicx}
\usepackage{multirow}
\usepackage{array}
\usepackage{siunitx}
\DeclareMathOperator*{\argmin}{arg\,min}
\DeclareMathOperator{\Tr}{Tr}

\usepackage{wrapfig}
\modulolinenumbers[5]
\usepackage{mathtools}

\begin{document}
\title{Integrating~Neural~Networks~and~Dictionary Learning~for~Multidimensional~Clinical~Characterizations~from~Functional~Connectomics~Data}
%
%
\authorrunning{Niharika Shimona D'Souza et al.} 
%
%
\author{Niharika Shimona D'Souza\inst{1} 
 \textsuperscript{*}
\and Mary Beth Nebel \inst{2} \inst{3}  \and Nicholas Wymbs \inst{2} \inst{3} \and Stewart Mostofsky \inst{2}\inst{3} \inst{4} \and Archana Venkataraman\inst{1} }


\institute{Dept. of Electrical and Computer Eng., Johns Hopkins University, Baltimore, USA
\email{\textsuperscript{*}}{Shimona.Niharika.Dsouza@jhu.edu}
\and
Center for Neurodevelopmental \& Imaging Research, Kennedy Krieger Institute
\and
Dept. of Neurology, Johns Hopkins School of Medicine, Baltimore, USA
\and
Dept. of Psychiatry and Behavioral Science, Johns Hopkins School of Medicine, USA}
\maketitle              
\begin{abstract}
We propose a unified optimization framework that combines neural networks with dictionary learning to model complex interactions between resting state functional MRI and behavioral data. The dictionary learning objective decomposes patient correlation matrices into a collection of shared basis networks and subject-specific loadings. These subject-specific features are simultaneously input into a neural network that predicts multidimensional clinical information. Our novel optimization framework combines the gradient information from the neural network with that of a conventional matrix factorization objective. This procedure collectively estimates the basis networks, subject loadings, and neural network weights most informative of clinical severity.  We evaluate our combined model on a multi-score prediction task using $52$ patients diagnosed with Autism Spectrum Disorder (ASD). Our integrated framework outperforms state-of-the-art methods in a ten-fold cross validated setting to predict three different measures of clinical severity. 
\end{abstract}

\section{Introduction}
Resting state fMRI (rs-fMRI) tracks steady-state co-activation patterns i.e. functional connectivity, in the brain in the absence of a task paradigm. Recently, there has been an increasing interest in using rs-fMRI to study neurodevelopmental disorders, such as autism and schizophrenia. These disorders are characterized by immense patient heterogeneity in terms of their clinical manifestations. However, the high data dimensionality coupled with external noise greatly confound a unified characterization of behavior and connectivity data.
\par Most techniques relating rs-fMRI to behavior focus on discriminating patients from controls or localising functional regions in the brain~\cite{nandakumar2018defining}. In the simplest case, statistical tests are used to identify group differences in rs-fMRI features \cite{vieira2017using}. From the machine learning side, neural network architectures have become popular for investigating neuroimaging correlates of developmental disorders \cite{vieira2017using}. However, very few works handle continuous valued severity prediction from connectomics data. One recent example is the work of \cite{kawahara2017brainnetcnn}, which develops a convolutional neural network (CNN) to predict two cognitive measures directly from brain connectomes. A more traditional example is the work of \cite{d2018generative}, which combines a dictionary learning on patient correlation matrices with a linear regression on the patient loadings to predict clinical severity. Their joint optimization procedure helps the authors extract representations that generalize to unseen data. A more pipelined approach is presented in \cite{rahim2017joint}. They decouple feature selection from prediction to estimate multiple severity measures jointly from voxel-ROI correlations. In contrast, our method \textit{jointly optimizes} for an \textit{anatomically interpretable basis} and a complex \textit{non-linear behavioral encoding} that explain connectivity and clinical severity simultaneously.
\par We propose one of the first end-to-end frameworks that embeds a traditional model-based representation (dictionary learning) with deep networks into a single optimization. We borrow from the work of \cite{d2018generative,d2019coupled} to project the patient correlation matrices onto a shared basis. However, in a notable departure from prior work, we couple the patient projection onto the dictionary with a neural network for multi-score behavioral prediction. We \textit{jointly optimize for the basis, patient representation, and neural network weights} by combining gradient information from the two objectives. We demonstrate that our unified framework provides us with the necessary representational flexibility to model complex interactions in the brain, and to learn effectively from limited training data. Our unique optimization strategy outperforms state-of-the-art baseline methods at estimating a generalizable multi-dimensional patient characterization.

\section{Multidimensional Clinical Characterization from Functional Connectomics}

\begin{figure}[t!]
   \centering
   \includegraphics[width=\dimexpr \textwidth-6\fboxsep-6\fboxrule\relax]{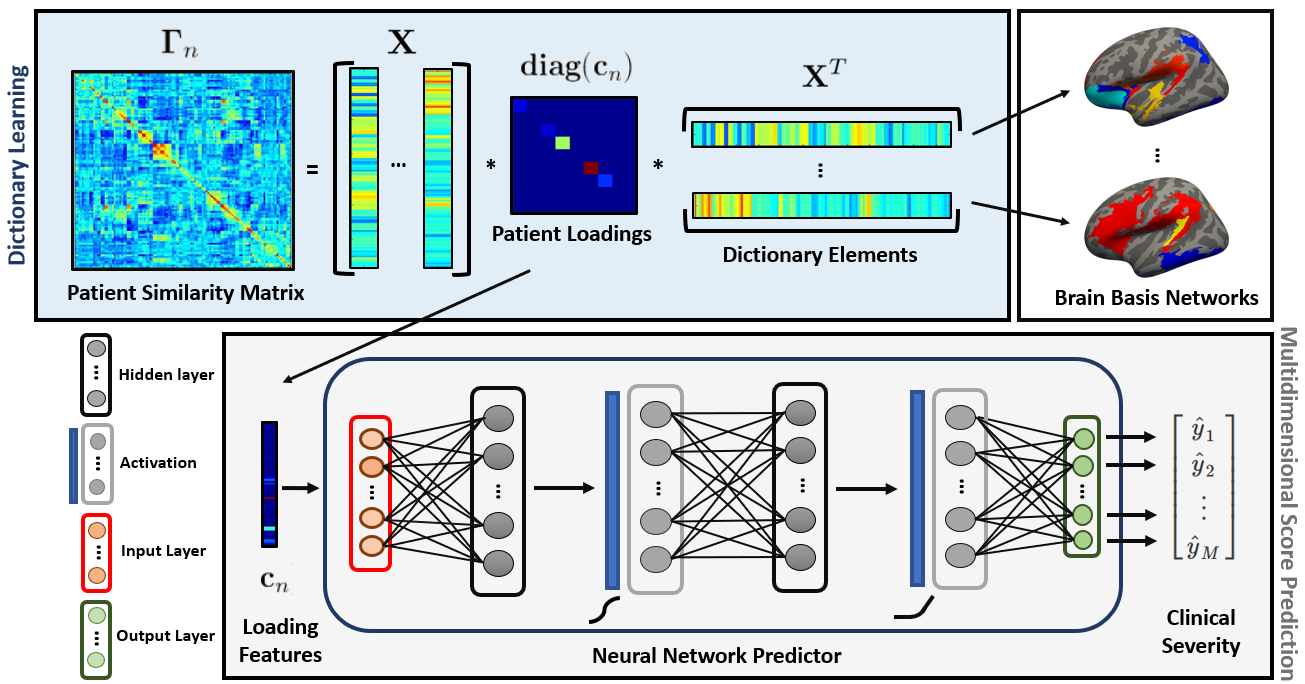}
   \small{\caption{\footnotesize{A unified framework for integrating neural networks and dictionary learning. \textbf{Blue Box:}~Dictionary Learning from correlation matrices \textbf{Gray Box:} Neural Network architecture for multidimensional score prediction} } \label{CCS}}
\end{figure}

Fig.~\ref{CCS} illustrates our framework. The blue box denotes our dictionary learning representation, while the gray box is the neural network architecture. Let $N$ be the number of patients and $P$ be the number of regions in our brain parcellation. We decompose the correlation matrix $\mathbf{\Gamma}_{n} \in \mathcal{R}^{P\times P} $ for each patient~$n$, via $K$ dictionary elements of a shared basis $\mathbf{X} \in \mathcal{R}^{P \times K}$, and a subject-specific loading vector  $\mathbf{c}_{n} \in \mathcal{R}^{K \times 1}$. Thus, our dictionary learning objective $\mathcal{D}$ is as follows:
\begin{equation}
\mathcal{D}(\mathbf{X},\{\mathbf{c}_{n}\};\{\mathbf{\Gamma}_{n}\}) = \sum_{n}\left[\vert\vert{\mathbf{\Gamma}_{n} - \mathbf{X}\mathbf{diag}(\mathbf{c}_{n}) \mathbf{X}^{T}\vert\vert}_{F}^{2} + \gamma_{2}{\vert\vert{\mathbf{c}_{n}\vert\vert}}^{2}_{2}\right] + \gamma_{1}{\vert\vert{\mathbf{X}}\vert\vert}_{1} 
\label{DictL}
\end{equation}
where $\mathbf{diag}(\mathbf{c}_{n})$ denotes a matrix with the entries of $\mathbf{c}_{n}$ on the leading diagonal and the non-diagonal entries as $0$. Since $\mathbf{\Gamma}_{n}$ is positive semi-definite, we add the constraint $\mathbf{c}_{nk} \geq 0$. The columns of $\mathbf{X}$ capture representative patterns of co-activation common to the cohort. The loadings $\mathbf{c}_{nk}$ capture the network strength of basis $k$ in patient $n$. We add an $\ell_{1}$ penalty to $\mathbf{X}$ to encourage sparsity, and an $\ell_{2}$ penalty to $\{\mathbf{c}_{n}\}$ to ensure that the objective is well posed. $\gamma_{1}$ and $\gamma_{2}$ are the corresponding regularization weights. 
The loadings $\mathbf{c}_{n}$ are also the input features to a neural network. The network parameters $\mathbf{\Theta}$ encode a series of non-linear transformations that map the input features to behavior. $\mathbf{Y}_{n} \in \mathcal{R}^{M \times 1}$ is a vector of $M$ concatenated clinical measures, which describe the location of patient~$n$ on the behavioral spectrum. $\mathbf{\hat{Y}}_{n}$ is estimated using the latent representation~$\mathbf{c}_{n}$. We employ the Mean Square Error (MSE) to define the network loss $\mathcal{L}$:
\begin{equation}
\mathcal{L}(\{\mathbf{c}_{n}\},\mathbf{\Theta};\{\mathbf{Y}_{n}\}) = \sum_{n}{\ell_{\mathbf{\Theta}}(\mathbf{c}_{n},\mathbf{Y}_{n})} = \lambda \sum_{n}{\vert\vert{\mathbf{\hat{Y}}_{n}-\mathbf{{Y}}_{n}}\vert\vert}^{2}_{F}
\label{MSE}    
\end{equation}
Since $\mathcal{L}$ is added to $\mathcal{D}$ defined in Eq.~(\ref{DictL}), $\lambda$ balances the contribution of the dictionary learning and neural network terms to the objective.
\par Our proposed network architecture is highlighted in the gray box. Our modeling choices require us to carefully control for two key network design aspects: representational capacity, and convergence of the optimization. Given the low dimensionality of the input $\mathbf{c}_{n}$, we opt for a simple fully connected Artificial Neural Network (ANN) with two hidden layers and a width of $40$. We use a tanh function $\Big(\tanh(x) = \frac{\exp(x)- \exp(-x)}{\exp(x)+\exp(-x)}\Big)$ as the activation to the first hidden layer. We then use a SoftPlus ($SP(x) = \log(1+ \exp(x))$), a smooth approximation to the Rectified Linear Unit, as the activation for the second layer. Experimentally, we found that these modeling choices are robust to issues with saturation and vanishing gradients, which commonly confound neural network training.

\subsection{Joint Optimization Strategy:}

We use alternating minimization to iteratively optimize for the dictionary elements $\mathbf{X}$, the patient projections $\{\mathbf{c}_{n}\}$, and ANN parameters $\mathbf{\Theta}$. Here, we sequentially optimize for each hidden variable in the objective by fixing the rest, until global convergence. We use Proximal Gradient Descent to handle the non-differentiable $\ell_{1}$ penalty in Eq.~(\ref{DictL}), which requires the rest of the objective to be convex in $\mathbf{X}$. We circumvent this issue by the strategy in \cite{d2018generative}. Namely, we introduce $N$ constraints of the form $\mathbf{V}_{n} = \mathbf{X}\mathbf{diag}(\mathbf{c}_{n})$, and substitute them into the Frobenious norm terms in Eq.~(\ref{DictL}). These constraints are enforced using the Augmented Lagrangians $\{\mathbf{\Lambda}_{n}\}$. If $\Tr[\mathbf{Q}]$ denotes the trace operation, we add $N$ terms of the form ${\Tr{\left[{\mathbf{\Lambda}_{n}^{T}({\mathbf{V}_{n}-\mathbf{X}\mathbf{diag}(\mathbf{c}_{n})})}\right]}}+{{0.5}~{\vert\vert{\mathbf{V}_{n}-\mathbf{X}\mathbf{diag}(\mathbf{c}_{n})}\vert\vert}_{F}^{2}}$ to Eq.~(\ref{DictL}). We then iterate through the following four steps until convergence.
\paragraph{\textbf{Proximal Gradient Descent on $\mathbf{X}$.}}
Each step of the proximal algorithm constructs a a locally smooth quadratic model of $\vert\vert{\mathbf{X}}\vert\vert_{1}$ based on the gradient of $\mathcal{D}$ with respect to $\mathbf{X}$. Using this model, the algorithm iteratively updates $\mathbf{X}$ through shrinkage thresholding. We fix the learning rate for this step at $10^{-4}$.
\paragraph{\textbf{Updating the Neural Network Weights $\mathbf{\Theta}$.}} We optimize the weights $\mathbf{\Theta}$ according to the loss function $\mathcal{L}$ using backpropagation to estimate gradients. There are several obstacles in training a neural network to generalize and few available theoretical guarantees to guide design considerations. We pay careful attention to this, since the global optimization procedure couples the updates between $\mathbf{\Theta}$ and $\{\mathbf{c}_{n}\}$. We employ the ADAM optimizer, which is robust to small datasets. We randomly initialize at the first main update. We found a learning rate of $10^{-4}$, scaled by $0.9$ every $5$ epochs to be sufficient for encoding the training data, while avoiding bad local minima and over-fitting. We train for $50$ epochs with a batch-size of $12$. Finally, we fix the obtained weights to update  $\{\mathbf{c}_{n}\}$.
\paragraph{\textbf{L-BFGS update for $\{\mathbf{c}_{n}\}$.}}
The objective for each $\mathbf{c}_{n}$ decouples as follows:
\begin{multline}
\mathcal{J}(\mathbf{c}_{n}) = \ell_{\mathbf{\Theta}}(\mathbf{c}_{n},\mathbf{Y}_{n}) + \gamma_{2}{\vert\vert{\mathbf{c}_{n}}\vert\vert}^{2}_{2} + {\Tr\left[{\mathbf{\Lambda}^{T}_{n}}({\mathbf{V}_{n}-\mathbf{X}\mathbf{diag}(\mathbf{c}_{n}))}\right]} \\ +    
{0.5} {\vert\vert{\mathbf{V}_{n}-\mathbf{X}\mathbf{diag}(\mathbf{c}_{n})}\vert\vert}^{2}_{F} \ \ \ \ s.t. \ \ \  \mathbf{c}_{nk} \geq 0
\label{lbfgs}
\end{multline}
Notice that we can use a standard backpropagation algorithm to compute the gradient of $\ell_{\mathbf{\Theta}}(.)$ with respect to $\mathbf{c}_{n}$, denoted by $\nabla{\ell_{\mathbf{\Theta}}(\mathbf{c}_{n},\mathbf{Y}_{n})}$. The gradient of $\mathcal{J}$ with respect to $\mathbf{c}_{n}$, denoted $\mathbf{g}(\mathbf{c}_{n})$, can then be computed as follows:
\begin{equation*}
\mathbf{g}(\mathbf{c}_{n}) = \nabla{\ell_{\mathbf{\Theta}}(\mathbf{c}_{n},\mathbf{Y}_{n})} + {\mathbf{c}_{n} \circ \left[\left[\mathcal{I}_{K}\circ(\mathbf{X}^{T}\mathbf{X})\right]\mathbf{1}\right]} - \left[\mathcal{I}_{K}\circ(\mathbf{\Lambda}_{n}^{T}\mathbf{X}+\mathbf{V}_{n}^{T}\mathbf{X})\right]\mathbf{1} + 2\gamma_{2}\mathbf{c}_{n}
 \label{grad}
\end{equation*}
where $\mathbf{1}$ is the vector of all ones, and $\circ$ represents the Hadamard product. The first term is from the ANN, while the rest are from the modified dictionary learning objective. The gradient combines information from the ANN function landscape with that from the correlation matrix estimation. For each iteration~$r$, the BFGS \cite{wright1999numerical} algorithm recursively constructs a positive-definite Hessian approximation $\mathbf{B}(\mathbf{c}^{r}_{n})$ based on the gradients estimated. Next, we iteratively compute a descent direction $\mathbf{p}$ for $\mathbf{c}^{r}_{n}$ using the following bound-constrained objective:
\begin{equation}
    \mathbf{p}^{*} = \argmin_{\mathbf{p}}{\mathcal{J}(\mathbf{c}_{n}^{r})+{\mathbf{g}(\mathbf{c}_{n}^{r})^{T}}{\mathbf{p}} + {0.5} {\mathbf{p}^{T}\mathbf{B}(\mathbf{c}_{n}^{r}){\mathbf{p}}}} \ \ s.t. \  \ \mathbf{c}^{r}_{nk} + \mathbf{p}_{k} \geq 0
    \label{BFGS}
\end{equation}
We then update $\mathbf{c}_{n}$ as: $\mathbf{c}^{r+1}_{n} = \mathbf{c}^{r}_{n} + \delta \mathbf{p}^{*}$, repeating this procedure until convergence. Effectively, the BFGS update leverages second-order curvature information through each $\mathbf{B}(\mathbf{c}_{n})$ estimation. In practice, $\delta$ is set to $0.9$.
\paragraph{\textbf{Augmented Lagrangian Update for the Constraint Variables.}}
We have a closed form solution for computing the constraint argument $\{\mathbf{V}_{n}\}$. The  dual Lagrangians, i.e. $\{\mathbf{\Lambda}_{n}\}$ are updated via gradient ascent. We cycle through the collective updates for these two variables until convergence. We use a learning rate of $10^{-4}$, scaled by $0.75$ at each iteration of gradient ascent.
\paragraph{\textbf{Prediction on Unseen Data.}}
We use cross validation to assess our framework. For a new patient, we compute the loading vector $\mathbf{\Bar{c}}$ using the estimates  $\{\mathbf{X}^{*},\mathbf{\Theta}^{*}\}$ obtained during training. We remove the contribution of the ANN term from the joint objective, as we do not know the corresponding value of $\mathbf{\Bar{Y}}$ for a new patient. The proximal operator conditions are assumed to hold with equality, removing the Lagrangian terms. The optimization in $\mathbf{\Bar{c}}$ takes the following form:
\begin{align}
{0.5}~{\mathbf{\Bar{c}}^{T}\mathbf{\Bar{H}}\mathbf{\Bar{c}}} + \mathbf{\Bar{f}}^{T}\mathbf{\Bar{c}} \ \ s.t. \ \ \mathbf{\Bar{A}}\mathbf{\Bar{c}} \leq \mathbf{\Bar{b}} \ \ \ \ \ \ \ \  \label{qp} \\  \notag
\mathbf{\Bar{H}} = 2(\mathbf{X}^{T}\mathbf{X})\circ(\mathbf{X}^{T}\mathbf{X}) + 2\gamma_{2}\mathcal{I}_{K} \ \ \ \ \\  \notag 
 \mathbf{\Bar{f}} = -2\mathcal{I}_{K}\circ(\mathbf{X}^{T}\mathbf{\Gamma}_{n}\mathbf{X})\mathbf{1}; \ \  \mathbf{\Bar{A}} = -\mathcal{I}_{K} \ \  \mathbf{\Bar{b}} = \mathbf{0} 
\end{align}
This formulation is similar to Eq.~(\ref{BFGS}) from the BFGS update for $\{\mathbf{c}_{n}\}$. $\mathbf{\Bar{H}}$ is also positive definite, thus giving an efficient quadratic programming solution to Eq.~(\ref{qp}). We estimate the score vector $\bar{\mathbf{Y}}$ by a forward pass through the ANN. 
\subsection{Baseline Comparisons}
We compare against two baselines that predict severity scores from correlation matrices $\mathbf{\Gamma}_{n} \in \mathcal{R}^{P \times P}$. The first has a joint optimization flavor similar to our work, while the second uses a CNN to exploit the structure in $\{\mathbf{\Gamma}_{n}\}$:
\begin{itemize}
\item[1.]{The Generative-Discriminative Basis Learning framework in \cite{d2018generative} }
\item[2.]{BrainNet Convolutional Neural Network (CNN) from \cite{kawahara2017brainnetcnn} }
\end{itemize}
\paragraph{\textbf{Implementation Details.}} The model in \cite{d2018generative} adds a linear predictive term  $\gamma{\vert\vert{\mathbf{C}^{T}\mathbf{w}-\mathbf{y}}\vert\vert}^{2}_{2}+ \lambda_{3}\vert\vert{\mathbf{\mathbf{w}}}\vert\vert_{2}^{2}$ to the dictionary learning objective~in~Eq.(\ref{DictL}). They estimate a single regression vector $\mathbf{w}$ to compute a scalar measure $\mathbf{y}_{n}$ from the loading matrix $\mathbf{C} \in \mathcal{R}^{K\times N}$. To provide a fair comparison, we modify this discriminative term to $\gamma{\vert\vert{\mathbf{C}^{T}\mathbf{W}-\mathbf{Y}}\vert\vert}^{2}_{2}+ \lambda_{3}\vert\vert{\mathbf{\mathbf{W}}}\vert\vert_{2}^{2}$, to predict the vectors $\{\mathbf{Y}_{n} \in \mathcal{R}^{M \times 1}\}_{n=1}^{N}$ using the weight matrix $\mathbf{W} \in \mathcal{R}^{K\times M}$. Using the guidelines in \cite{d2018generative}, we fixed $\lambda_{3}$ and $\gamma$ at $1$, and swept the other parameters over a suitable range. We set number of networks to $K=8$, which is the knee point of the eigenspectrum for $\{\mathbf{\Gamma}_{n}\}$.
\par The network architecture in \cite{kawahara2017brainnetcnn} predicts two cognitive measures from correlation matrices. In our case, $\{\mathbf{\Gamma}_{n}\}$ are of size $P \times P$. For our comparison, we modify the output layer to be of size $M$. We use the recommended guidelines in \cite{kawahara2017brainnetcnn} for setting the learning rate, batch-size and momentum during training.   
\par For our framework, the trade-off $\lambda$ from Eq.~(\ref{MSE}) balances the dictionary learning and neural network losses in the joint optimization. The generalization is also governed by $\gamma_{1}$ and $\gamma_{2}$ from the dictionary learning. Using a grid search, we fix $\{\gamma_{1} = 10, \gamma_{2} = 0.1, \lambda = 0.1 \}$. The number of networks $K$ is fixed to $8$.
\section{Experimental Evaluation and Results:}
\paragraph{\textbf{Data and Preprocessing.}} We validate our method on a cohort of $52$ children with high-functioning ASD. Rs-fMRI data is acquired  on a Phillips $3$T Achieva scanner (EPI, with TR/TE $=2500/30$ms, flip angle $=70^{\circ}$, res $=3.05\times3.15\times3$mm, having $128$ or $156$ time samples). We use the pre-processing pipeline in \cite{d2018generative}, which consists of slice time correction, rigid body realignment, normalization to the EPI version of the MNI template, Comp Corr, nuisance regression, spatial smoothing by a $6$mm FWHM Gaussian kernel, and bandpass filtering ($0.01-0.1$Hz). We defined $116$ regions using the Automatic Anatomical Labeling (AAL) atlas. The contribution of the first eigenvector is subtracted from the regionwise correlation matrices because it is roughly constant and biases the predictions. The residual correlation matrices, $\{\mathbf{\Gamma}_{n}\}$, are used as inputs for all three methods.
\par We use three clinical measures quantifying various impairments associated with ASD. The Autism Diagnostic Observation Schedule (ADOS) \cite{payakachat2012autism} is scored by a clinician and captures socio-communicative deficits along with repetitive behaviors (dynamic range: $0-30$). The Social Responsiveness Scale (SRS) \cite{payakachat2012autism} is scored through a parent/teacher questionnaire, and quantifies impaired social functioning (dynamic range: $70-200$). On the other hand, Praxis measures the ability to perform skilled motor gestures on command, by imitation, and for tool usage. Two trained research-reliable raters score a videotaped performance based on total percent correct gestures (dynamic~range:~$0-100$). 
\begin{figure}[t!]    
    \centering
      \includegraphics[width=\dimexpr \textwidth-2\fboxsep-2\fboxrule\relax]{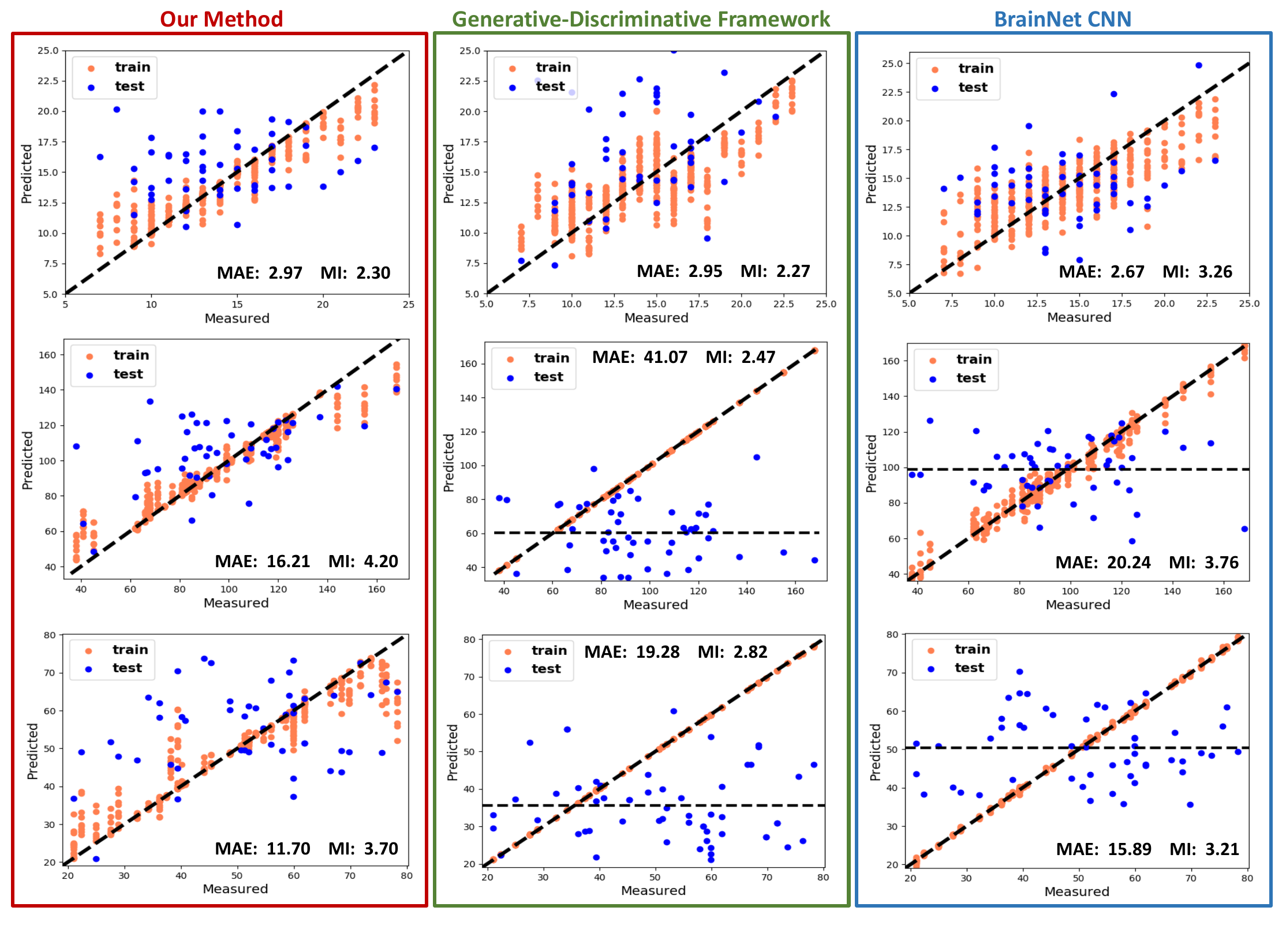}
   \small{\caption{Multi-Score Prediction performance for \textbf{Top:} ADOS \textbf{Middle:} SRS \textbf{Bottom:} Praxis by \textbf{Red Box:} Our Framework. \textbf{Green Box:} Generative-Discriminative Framework from \cite{d2018generative}. \textbf{Blue Box:} BrainNet CNN from \cite{kawahara2017brainnetcnn}}}
   
   \label{Multi-Score}
\end{figure}%
\paragraph{\textbf{Performance Characterization.}} Fig.~\ref{Multi-Score} illustrates the \textit{multi-score regression} performance of each method based on ten fold cross validation. Our quantitative metrics are median absolute error (MAE) and mutual information (MI) between the actual and computed scores. Lower MAE and higher MI indicate better performance. The orange points indicate training fit, while the blue points denote performance on held out samples. The $\mathbf{x} = \mathbf{y} $ line indicates ideal performance.
\par Observe that both the Generative-Discriminative model and the BrainNet CNN perform comparably to our model for predicting ADOS. However, our model outperforms the baselines in terms of MAE and MI for SRS and Praxis, with the blue points following the $\mathbf{x}=\mathbf{y}$ line more closely. Generally, we find that as we vary the free parameters, the baselines predict \textit{one of the three scores well (in Fig.~\ref{Multi-Score}, ADOS), but fit the rest poorly}. In contrast, only our framework learns a representation that predicts all three clinical measures \textit{simultaneously}, and hence overall outperforms the baselines. We believe that the representational flexibility of neural networks along with our joint optimization is key to generalization.
\par
\begin{figure}[t!]
 \centerline{\includegraphics[scale=0.25]{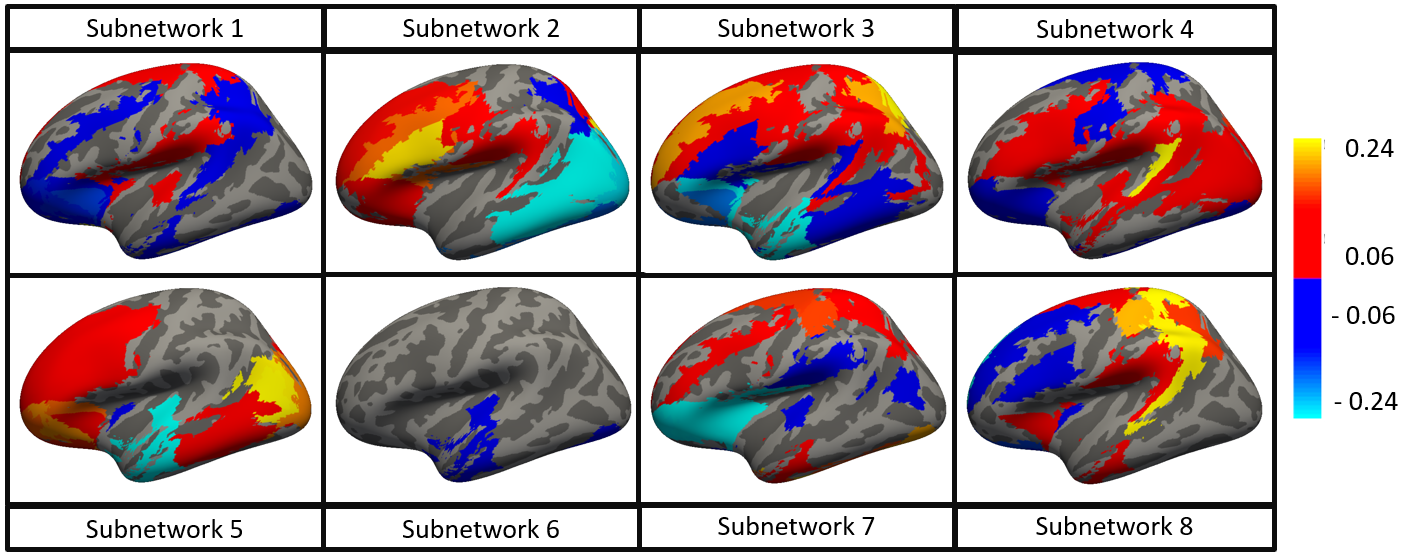}}
\small{{\caption{\footnotesize{Eight subnetworks identified by our model from multi-score prediction. The blue and green regions are anticorrelated with the red and orange regions.}}} \label{ADOS_NL}}
\end{figure}

 Fig.~\ref{ADOS_NL} illustrates the subnetworks in $\{\mathbf{X}_{k}\}$. Regions storing positive values are anticorrelated with negative regions. From a clinical standpoint, Subnetwork~$8$ includes the somatomotor network (SMN) and competing, i.e. anticorrelated, contributions from the default mode network (DMN). Subnetwork~$3$ also has contributions from the DMN and SMN, both of which have been widely reported in ASD \cite{nebel2016intrinsic}. Along with the DMN, Subnetworks~$5$ and $2$ contain positive and competing contributions from the higher order visual processing areas (i.e. occipital and temporal lobes) respectively. These findings concur with behavioral reports of reduced visual-motor integration in ASD \cite{nebel2016intrinsic}. Finally, Subnetworks~$2$, $3$, and $8$ exhibit central executive control network and insula contributions, believed to be critical for switching between self-referential and goal-directed behavior \cite{sridharan2008critical}.
\section{Conclusion}
We have introduced the first unified framework to combine classical optimization with the modern-day representational power of neural networks. This integrated strategy allows us to characterize and predict multidimensional behavioral severity from rs-fMRI connectomics data. Namely, our dictionary learning term provides us with interpretability in the brain basis for clinical impairments. Our predictive term cleverly exploits the ability of neural networks to learn rich representations from data. The joint optimization procedure helps learn informative connectivity patterns from limited training data. Our framework makes very few assumptions about the data and can be adapted to work with complex clinical score prediction scenarios. For example, we are developing an extension our method to handle case/control severity prediction using a mixture density network (MDN) \cite{bishop1994mixture} in lieu of a regression network. The MDN models a mixture of Gaussians to fit the target bimodal distibution. Accordingly, the network loss function is a negative log-likelihood, which differs from conventional formulations. This is another scenario that may advance our understanding of neuropsychiatric disorders. In the future, we will also explore extensions that simultaneously integrate functional, structural and dynamic information.
{\paragraph{\textbf{Acknowledgements.}}
This work was supported by the National Science Foundation CRCNS award 1822575, National Science Foundation CAREER award 1845430, National Institute of Mental Health (R01 MH085328-09, R01 MH078160-07, K01 MH109766 and R01 MH106564), National Institute of Neurological Disorders and Stroke (R01NS048527-08), and the Autism Speaks foundation.}
\bibliographystyle{splncs04}
\small{{\bibliography{MyRefs.bib}}}
\end{document}